\pdfoutput=1

\documentclass[11pt]{article}

\usepackage{ACL2023}

\usepackage{times}
\usepackage{latexsym}
\usepackage{amsmath}
\usepackage{hyperref}

\usepackage[T1]{fontenc}
\usepackage{graphicx} 
\usepackage[utf8]{inputenc}

\usepackage{tabularx}
\usepackage{booktabs}
\usepackage{microtype}
\usepackage{longtable}
\usepackage{wrapfig}
\usepackage[utf8]{inputenc}
\usepackage{natbib}  

\usepackage{inconsolata}

\title{Decoding Climate Disagreement: A Graph Neural Network-Based Approach to Understanding Social Media Dynamics}

\author{Ruiran Su \\
  Department of Engineering Science \\
  University of Oxford \\
  \texttt{ruiran.su@trinity.ox.ac.uk} \\\And
  Janet B. Pierrehumbert \\
  Oxford e-Research Centre \\
  University of Oxford \\
  \texttt{janet.pierrehumbert@oerc.ox.ac.uk} \\}

\begin{document}
\maketitle
\begin{abstract}
This work introduces the ClimateSent-GAT Model, an innovative method that integrates Graph Attention Networks (GATs) with techniques from natural language processing to accurately identify and predict disagreements within Reddit comment-reply pairs. Our model classifies disagreements into three categories: agree, disagree, and neutral. Leveraging the inherent graph structure of Reddit comment-reply pairs, the model significantly outperforms existing benchmarks by capturing complex interaction patterns and sentiment dynamics. This research advances graph-based NLP methodologies and provides actionable insights for policymakers and educators in climate science communication.

\end{abstract}

\section{Introduction}
The urgency of addressing climate change is paralleled by the complexity of discussions it evokes on social media platforms. These platforms, functioning as contemporary public squares, host diverse opinions intertwined with misinformation \cite{Diggelmann2020}, posing significant challenges for distinguishing constructive debates from misleading discourse \cite{Johansson2023}. Traditional natural language processing (NLP) techniques often fall short in effectively understanding disagreements that characterize online discussions. 

Graph Neural Networks (GNNs), particularly Graph Attention Networks (GATs) \cite{Veličković2018}, have emerged as potent tools for modeling relational data in complex networks. Their capability to learn and represent relationships in data makes them ideally suited for modelling social media interactions, where the structure of dialogue can be as informative as the content itself. 

Thus, we present the ClimateSent-GAT Model in this paper, which not only exploits the textual and sentimental content of communications but also captures the intricate interactions of climate discourse on social media.
In the context of climate change discussions on platforms like Reddit, where comment-reply pairs form a natural graph structure, our model innovatively applies GATs to this domain. By focusing on the detection of disagreement in climate-related discourse, ClimateSent-GAT aims to shed light on the patterns of communication that propagate misinformation and foster contention. The objective is twofold: to advance the methodologies of NLP by integrating them with graph-based models, and to provide actionable insights that can aid policymakers, educators, and social media platforms in fostering a more informed and rational public discourse on climate change.
\begin{table}[h]
\centering
\begin{tabular}{|l|r|}
\hline
\textbf{Statistics} & \textbf{r/climate} \\ \hline
Start Date         & January 2015       \\ \hline
End Date         & May 2021       \\ \hline
Number of Posts   & 2367              \\ \hline
Number of Users    & 4,580              \\ \hline
Comment-Reply Interactions    & 5,773              \\ \hline
Interactions labelled as Agree     & 32\%               \\ \hline
Interactions labelled as Neutral      & 28\%               \\ \hline
Interactions labelled as Disagree     & 40\%               \\ \hline
\end{tabular}
\caption{Dataset statistics for the r/climate subreddit.}
\label{tab:climate_data}
\end{table}

Our study employed the Climate subset from the DEBAGREEMENT dataset, as described in \cite{PougueBiyong2021}. The DEBAGREEMENT dataset was constructed by harvesting data from various subreddits using the PushShift API, which offers historical data for research purposes. To ensure the dataset only included meaningful interactions, submissions and comments with minimal engagement (fewer than a set threshold of comments or words) were excluded. This filtering aimed to focus on more impactful discussions. The resulting dataset comprised high-quality interactions that form a complex web of communication dynamics, annotated for (dis)agreement based on comment-reply contexts. The dataset is available under a Creative Commons Attribution 4.0 International License and can be accessed via \url{https://scale.com/open-datasets/oxford}. The subset we used was taken from the r/climate subreddit, a community dedicated to discussions on climate issues. Established in 2008, r/climate has grown to encompass 99,000 members. The Climate subset comprises all submissions and posts from Jan 2015 to May 2021. Each comment length ranging from 10 to 100 words, and for the DEBAGREEMENT dataset, comment-reply interactions were labelled by crowd-workers as ``agree'', ``disagree'', or ``neutral''.  We used this dataset for the same three-way classification task, to evaluate our model's capability to detect disagreements within climate change-related comment-reply pairs. We demonstrate superior performance compared to pre-existing models (see Table 2).

\section{Literature Review}

\subsection{Graph Neural Networks in Understanding Social Dynamics}

Graph Neural Networks (GNNs) have emerged as a powerful tool for understanding complex network structures.  One of the foundational works in applying GNNs to social networks is  \citet{Kipf2017}, who demonstrated how Graph Convolutional Networks (GCNs) could be used to classify nodes in citation networks. Building on this approach, researchers have adapted similar models to more complex social structures, such as user interactions on social media platforms \cite{Hamilton2017}. These studies show that GNNs can effectively model relationships and interactions, leading to improved performance in tasks like community detection and influence prediction.

Graph Attention Networks (GATs), introduced by \citet{Veličković2018}, further enhance this capability by incorporating attention mechanisms that weigh the influence of neighboring nodes. This feature is particularly useful in social media contexts, where the relevance and influence of a comment can vary significantly based on the interaction dynamics. For instance, \citet{AbuElHaija2019} leveraged GATs to predict the future state of users in dynamic social networks, effectively mapping how interactions influence user behavior over time. Moreover, GATs can be deployed to tackle more direct social issues. For example, Gao et al. \cite{Gao2022} used GATs to study the diffusion of information in online social networks, identifying key patterns that signify misinformation spread. This is directly relevant to fields like climate science, where misinformation can have significant real-world impacts.

Overall, the integration of GNNs into the analysis of social media dynamics offers a promising perspective for not only detecting and understanding social interactions but also for intervening in a timely manner to guide discussions towards more constructive outcomes. The ongoing development in this field suggests a growing potential for GNNs to contribute significantly to our understanding of digital communication landscapes, especially in contentious domains like climate science where the clarity and accuracy of information are paramount.

\subsection{Graph Neural Networks in NLP}

Graph Neural Networks (GNNs) continue to make significant strides in the field of Natural Language Processing (NLP), providing advanced solutions to complex problems where traditional methods fall short. The adaptability of GNNs to encode relationships within data makes them particularly effective for tasks involving rich contextual and relational information.

Advancements in semantic role labelling leverage GNNs to incorporate deep contextual embeddings. \citet{Marcheggiani2017} present a novel approach that integrates contextual information with GNNs, enhancing the model's ability for semantic labelling.

\citet{Wang2021} introduce a cross-lingual graph neural network that models syntactic and semantic relationships across languages. This framework significantly improves text classification accuracy, particularly for low-resource languages, by capturing and utilizing the inherent linguistic structures across different language families.

\citet{Ghosal2020} developed DialogueGCN, a graph convolutional network tailored for emotion recognition in conversations. This model recognizes and interprets the emotional dynamics in dialogues by structuring the dialogue as a graph where nodes represent utterances and edges define the interaction dynamics, leading to more nuanced and accurate emotion recognition.

Expanding the use of GNNs to document-level tasks,  \citet{Yasunaga2017} explored multi-document summarization through a graph-based approach. Their model, which constructs graphs representing relationships between sentences across documents, demonstrates improved performance in identifying key information and generating coherent summaries, showcasing the potential of GNNs to manage and synthesize information from multiple text sources.

These applications highlight the versatility and robustness of GNNs in tackling diverse NLP challenges. By effectively capturing and processing relational data, GNNs not only improve the performance of NLP systems but also open new avenues for research and development in the field.

Given the hierarchical and interconnected nature of social media threads, GNNs offer a promising avenue for disagreement detection, which we exploit in our ClimateSent-GAT model.

\subsection{Disagreement Detection and Modeling Social Interactions}

Disagreement detection in online discussions is a critical area of research in NLP that has seen substantial advancements with the incorporation of machine learning techniques, particularly GNNs. This section explores the latest methodologies for modeling social interactions and detecting disagreements, emphasizing the integration of sophisticated NLP tools with social network analysis.

Early works in disagreement detection leveraged techniques like sentiment analysis and opinion mining \cite{Pang2008}. With the maturation of the field, more nuanced techniques such as argumentation mining emerged \cite{Cabrio2017}. 

Recent studies have pushed the boundaries of disagreement detection by employing advanced machine learning frameworks that integrate GNNs with other deep learning techniques. Huang et al.  \cite{Huang2021} utilized a Recurrent Graph Neural Networks (RGNN)  to effectively identify disagreement in online forums. The RGNN model captures both the textual content and the relational dynamics between participants, leading to a more nuanced understanding of disagreement.

Climate science discussions, given their polarized nature, make the understanding of disagreement indispensable. The DEBAGREEMENT dataset and the Stance Embeddings Model by Pougué-Biyong et al. \cite{PougueBiyong2023} have laid the groundwork in this specific domain. 

These developments represent a leap forward in our ability to not only detect but also interpret and respond to disagreements in digital communication. However, most existing approaches have focused solely on textual features, missing out on the rich contextual cues available in the conversational structure of social media threads. 

\section{Experiments}

\subsection{Climate-related Entities Compilation}

This research uses the Climate subset of DEBAGREEMENT.  Our goal is to train a hybrid model which exploits both user interactions and sentiment features in the discourse towards climate-related entities. Thus, we firstly executed Named Entity Recognition (NER) using the SpaCy model. Then, we filtered out entities from non-relevant categories such as cardinal numbers, dates, and monetary values. The remained entities are still messy, so we manually compiled a climate-related entity list (see Appendix A) based on automatically extracted entities. We ended up having 1397 climate-related entities for further experiments.

\subsection{ClimateSent-GAT Model Construction}

In this paper, we introduce a hybrid model architecture that leverages textual embeddings, sentiment scores and Graph Attention Networks (GATs) to capture contextual and semantic information effectively. In the realm of climate science discussions, understanding the social relations and sentiment interactions towards climate entities in comment-reply structures can offer profound insights into public perceptions and discourse dynamics. Our ClimateSent-GAT model, a specialized variant of the Graph Attention Network, is tailored to capture these intricate sentiment relationships.

The choice of GATs over other types of GNNs such as Graph Convolutional Networks (GCNs) or Graph Recurrent Networks (GRNs) was motivated by several key considerations:

\begin{itemize}
    \item \textbf{Dynamic Edge Weighting:} 
    Unlike GCNs, which utilize fixed weights for edges based on the graph structure, GATs dynamically compute the weights through attention mechanisms. This adaptability is essential for social media, where the relevance of comments can significantly vary based on the context and interaction dynamics.
    
    \item \textbf{Fine-Grained Attention:} 
    GATs provide fine-grained control over information flow between nodes (e.g., comments and replies), focusing on the most informative parts of the data. This feature is crucial for environments like online forums, where not all interactions directly contribute to the outcomes of sentiment or disagreement detection.
    
    \item \textbf{Robustness to Sparse Data:} 
    Online discussions are often characterized by sparsity and uneven distribution. GATs excel in these settings by concentrating attention on significant nodes and edges, thus enhancing the model's predictive accuracy and reducing background noise.
    
    \item \textbf{Enhanced Feature Integration:} 
    The architecture of GATs allows for a nuanced integration of node and edge features, such as textual embeddings and sentiment scores. This integration is vital for detecting subtle cues that signify agreement or disagreement in communication.
\end{itemize}

In this section, we'll further introduce the details of the model.

\subsubsection{Feature Engineering}

We utilize a multi-faceted feature engineering approach that combines transformer-based sentence embeddings and sentiment scores to form a robust and contextually rich representation of social media dialogues.

To encode the textual features of the dataset, we utilize Sentence-BERT (paraphrase-MiniLM-L6-v2). The model generates 384-dimensional vectors that capture semantic meanings and syntactic structures for each sentence in the dialogue. The textual embeddings are then utilized as node features in our graph-structured data.

Sentiment analysis has been effectively applied to understand public opinion and user-generated content \cite{Cabrio2017}. To further enrich our feature set, we incorporate entity-based sentiment scores towards the climate-related entities for both parent and child messages in a conversation thread.   For each climate-related entity identified in the text, a snippet comprising 30 characters before and after the entity mention was extracted. The sentiment scores serve as an additional source of information, capturing the emotional tone and nuance in the dialogues, which is pivotal in discerning disagreement or agreement among users.

We firstly extracted the comment-reply pairs which mention at least one of the climate-related entities in the list, either in the parent text or the child text, based on the assumption that even if an entity isn't mentioned in both parent and child text but still is the subject of the discussion. We utilized HuggingFace's transformers library and initialized a sentiment analysis pipeline, this choice was motivated by the necessity to understand nuanced emotional expressions in social media texts. We designed a function which locates the mention of an entity and extracts a small context window around it (30 characters before and after the entity). The sentiment within this window is then evaluated, assigning a score and label based on the content's sentiment concerning the entity. If the entity is not mentioned in the text, a neutral sentiment is automatically assigned. This approach ensures that the sentiment analysis is focused and relevant to the specific topic being discussed rather than the entire comment, which may contain multiple sentiments.

We ended up gathering 8721 rows of comment-reply pairs with
sentiment-parent and sentiment-child scores towards each climate-related entity.

\subsubsection{Model Architecture}

The Climate subset of DEBAGREEMENT we use is inherently hierarchical and can be modeled effectively as a graph. In our graph representation, each node corresponds to a message in a thread of the social media interaction, with edges representing the parent-child relationship between messages.

Each parent-child comment pair forms two nodes in our graph. Specifically, a node corresponding to a parent comment will have the textual embedding of the parent comment and its sentiment score; another node corresponding to a child comment will have the textual embedding of the child comment and its sentiment score (see Fig.1). Edges between nodes are formed based on the parent-child relationships. A directed edge is created from the parent message to the child message, capturing the flow of social conversations.

Each node (comment) has features based on textual embeddings and sentiment scores. Differences in sentiment might be a straightforward indicator of potential disagreement.

For capturing topological and contextual information, we deploy Graph Attention Networks (GATs). Our architecture consists of two GAT layers:

\begin{itemize}
    \item The first GAT layer has 64 output channels with 8 attention heads. This layer is responsible for capturing local structural information.
    
    \item The second GAT layer further refines these features into 64 dimensions, serving to abstract higher-level features from the graph.

\end{itemize}
Both GAT layers use dropout for regularization and the Exponential Linear Unit (ELU) activation function for introducing non-linearity.

\begin{figure}[h]
    \centering
    \includegraphics[width=0.45\textwidth]{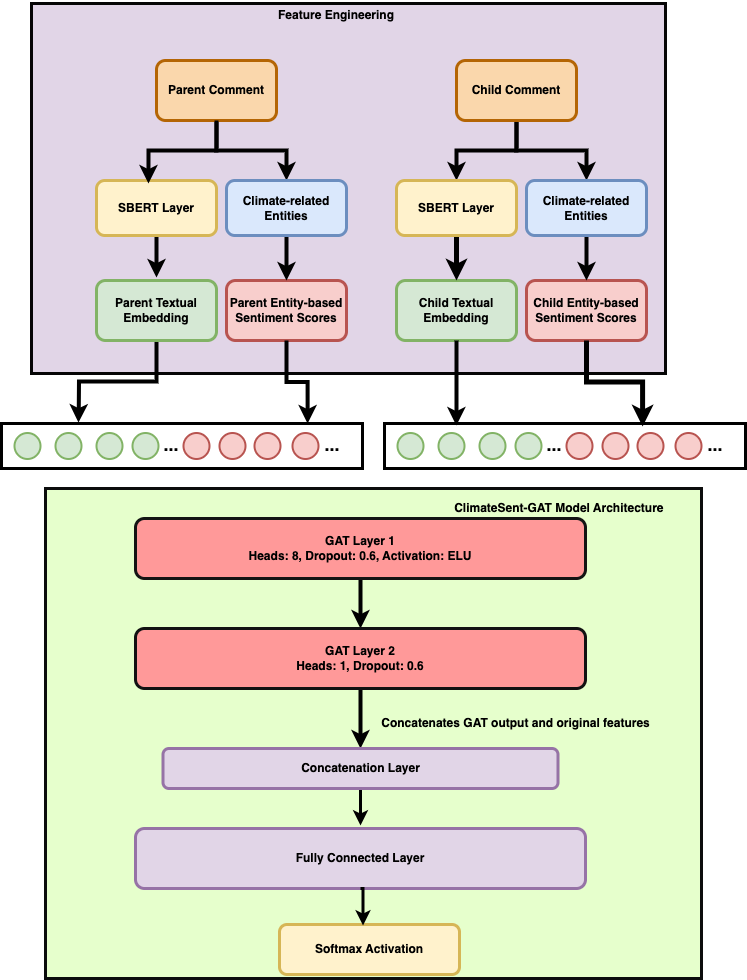}
    \caption{Diagram for the pipeline of the ClimateSent-GAT model}
\end{figure}

The core of ClimateSent-GAT lies in fusing graph-based features with textual and sentiment features. This captures both the contextual information within a thread and the semantic information of each individual message.

The concatenated feature vector is then passed through a fully connected layer that has three output units corresponding to our classes: Disagree (Class 0), Neutral (Class 1), and Agree (Class 2). A softmax activation is then applied to convert the logits into class probabilities.

Given a directed graph \( G = (V, E) \), where \( V \) denotes the set of nodes representing both parent and child comments in social media threads, and \( E \) represents the set of edges indicating reply relationships, we construct the graph's adjacency matrix and node features to train the ClimateSent-GAT model.

For a given pair of nodes \( i \) and \( j \), the raw attention coefficient \( e_{ij} \) is computed as:

\begin{align}
e_{ij} = \text{ELU} & \left( a^T \left[ W_1 h_{i, \text{embed}} \,\Vert\, W_2 h_{i, \text{sentiment}} \right. \right. \nonumber \\
& \left. \left. \,\Vert\, W_3 h_{j, \text{embed}} \,\Vert\, W_4 h_{j, \text{sentiment}} \right] \right)
\end{align}

Here, \( W_1, W_2, W_3, \) and \( W_4 \) are transformation matrices specific to different feature subsets (textual embeddings and sentiment scores for both parent and child comments). \( a^T \) is a transposed learnable weight vector. The ELU activation function ensures that the network maintains gradient flow even when negative attention coefficients are encountered.

To normalize the attention coefficients, we use:

\begin{equation}
\alpha_{ij} = \frac{\exp(e_{ij})}{\sum_{k \in N(i)} \exp(e_{ik})}
\end{equation}

Here, \( \alpha_{ij} \) is the normalized attention coefficient, and \( N(i) \) denotes the neighbors of node \( i \).

Our model integrates multi-head attention, computed as:

\begin{align}
h'_i = \Big\Vert_{k=1}^K \sigma & \left( \sum_{j \in N(i)} \alpha_{ij}^k \left[ W_{1}^k h_{j, \text{embed}} \right. \right. \nonumber \\
& \left. \left. \,\Vert\, W_{2}^k h_{j, \text{sentiment}} \right] \right)
\end{align}

Each updated node feature \( h'_i \) incorporates information from \( K \) different attention heads, each with their own transformed versions of the node features. This allows for more diverse and richer representations.

\subsubsection{Training and Evaluation}

We divided the dataset into training, validation, and testing subsets using a 70-15-15 percentage split, respectively. 

For reproducibility, we set up a fixed random seed of 42 for both NumPy and PyTorch. We trained the model using the Adam optimizer with a learning rate of \(0.001\) and a weight decay of \(5 \times 10^{-4}\). To address class imbalance, we also oversampled the minority class (Class 1, Neutral) to ensure a balanced representation of classes in the training process. The class weights are computed based on the oversampled dataset to further mitigate the imbalance issue during model training.

Additionally, to prevent overfitting, we implemented an early stopping mechanism. The patience for early stopping is set to 20 epochs. After training, we evaluated the model on the test set to assess its performance.

\begin{table*}[!htb]
\centering
\caption{Comparison of Classification Metrics for Different Models}
\label{tab:classification-metrics}
\begin{tabular}{lcccccc}
\hline
Metrics & ClimateSent-GAT & GAT & BERT & RoBERTa & ClimateBERT \\ 
\hline
\multicolumn{6}{l}{\textbf{Class 0 - Disagree}} \\ 
\hline
Precision & \textbf{0.87} & 0.50 & 0.50 & \underline{0.73} & \underline{0.73} \\
Recall & \underline{0.78} & 0.33 & \textbf{0.88} & 0.56 & 0.56 \\
F1-score & \textbf{0.82} & 0.39 & \underline{0.64} & 0.63 & 0.63 \\ 
\hline
\multicolumn{6}{l}{\textbf{Class 1 - Neutral}} \\ 
\hline
Precision & \textbf{0.65} & 0.16 & \underline{0.51} & 0.36 & 0.32 \\
Recall & \textbf{0.81} & 0.21 & 0.31 & \underline{0.69} & 0.67 \\
F1-score & \textbf{0.72} & 0.18 & 0.39 & \underline{0.48} & 0.44 \\ 
\hline
\multicolumn{6}{l}{\textbf{Class 2 - Agree}} \\ 
\hline
Precision & \textbf{0.78} & 0.36 & 0.58 & 0.60 & \underline{0.61} \\
Recall & \textbf{0.80} & 0.48 & 0.13 & \underline{0.57} & 0.54 \\
F1-score & \textbf{0.79} & 0.41 & 0.21 & \underline{0.58} & 0.57 \\ 
\hline
\multicolumn{6}{l}{\textbf{Overall Metrics}} \\ 
\hline
Macro Avg Precision & \textbf{0.76} & 0.34 & 0.53 & \underline{0.56} & \underline{0.56} \\
Macro Avg Recall & \textbf{0.80} & 0.34 & 0.44 & \underline{0.60} & 0.59 \\
Macro Avg F1-score & \textbf{0.78} & 0.33 & 0.41 & \underline{0.56} & 0.55 \\
Weighted Avg Precision & \textbf{0.80} & 0.39 & 0.53 & \underline{0.63} & \underline{0.63} \\
Weighted Avg Recall & \textbf{0.79} & 0.36 & 0.51 & \underline{0.58} & 0.57 \\
Weighted Avg F1-score & \textbf{0.80} & 0.37 & 0.44 & \underline{0.59} & 0.58 \\
Accuracy & \textbf{0.79} & 0.36 & 0.51 & \underline{0.58} & 0.57 \\ 
\hline
\end{tabular}
\end{table*}

The experiments in this research aim to predict labels for social media interactions through the proposed ClimateSent-GAT model, as showed in Table \ref{tab:classification-metrics}. For comparative purposes, we also run the standalone GAT model without any textual embeddings. For baseline models, we choose BERT uncased and RoBERTa base models. These are reported in \citet{PougueBiyong2021} as yielding F1 scores of 64.2\% and 63.3\%, respectively,  on the same classification task when averaged across all five DEBAGREEMENT subreddits. 
ClimateBert \cite{webersinke2021climatebert} was used to define a further baseline. We mainly focus on key metrics that best evaluate the model's performance based on our focus on predicting (dis)agreement among users on social media. 

For the `Disagree` class, we found out that ClimateSent-GAT model achieves the highest precision and F1-score across all models. The precision of \(0.87\) suggests that the model is reliable at identifying disagreeing comment pairs in climate-related discourse.   Interestingly, ClimateSent-GAT scores higher in recall (\(0.78\)), indicating that it might be more sensitive to capturing disagreement but at the cost of more false positives, which implies that it struggles to capture most of the disagreeing instances from the dataset. This is significant given that identifying disagreement is critical for dialog systems, sentiment analysis, and other NLP tasks related to social interactions on climate change.

Secondly, for the `Neutral` class, ClimateSent-GAT model again scores the highest in terms of precision and F1-score. Interestingly, ClimateSent-GAT scores higher in recall (\(0.79\)), indicating that it might be more sensitive to capturing neutral sentiments but at the cost of more false positives, as evidenced by the lower precision (\(0.71\)).

ClimateSent-GAT surpasses all models in all metrics for the `Agree` and 'Neutral' class. The model demonstrates its effectiveness at both accurately identifying and capturing most of the agreeing and neutral instances. 

Overall speaking, our model considerably outperforms the other models across all overall metrics. With macro-average and weighted-average F1-scores of 0.78 and 0.80 respectively, ClimateSent-GAT sets a new state-of-the-art for predicting disagreement between comment-reply pairs in climate change discussions.

\subsubsection{Improve the interpretability of existing model}
ClimateSent-GAT is a hybrid model incorporates both graph and text data, and the model is inherently complex, making it a good candidate for post-hoc interpretability methods.

The attention mechanism of our model assigns different weights to interactions in a graph. Thus, we first extract the attention weights from each layer of the trained model. These attention weights are then initially averaged across the heads for each layer to simplify the representation. Finally, we combine the averaged attention scores from both layers to obtain a single set of attention scores.  (see Figure 2) If the model learns to associate certain patterns of interaction (captured through embeddings and sentiment scores) with (dis)agreement, it will assign higher attention weights to such interactions. 
\begin{figure}[h]
    \centering
    \includegraphics[width=0.45\textwidth]{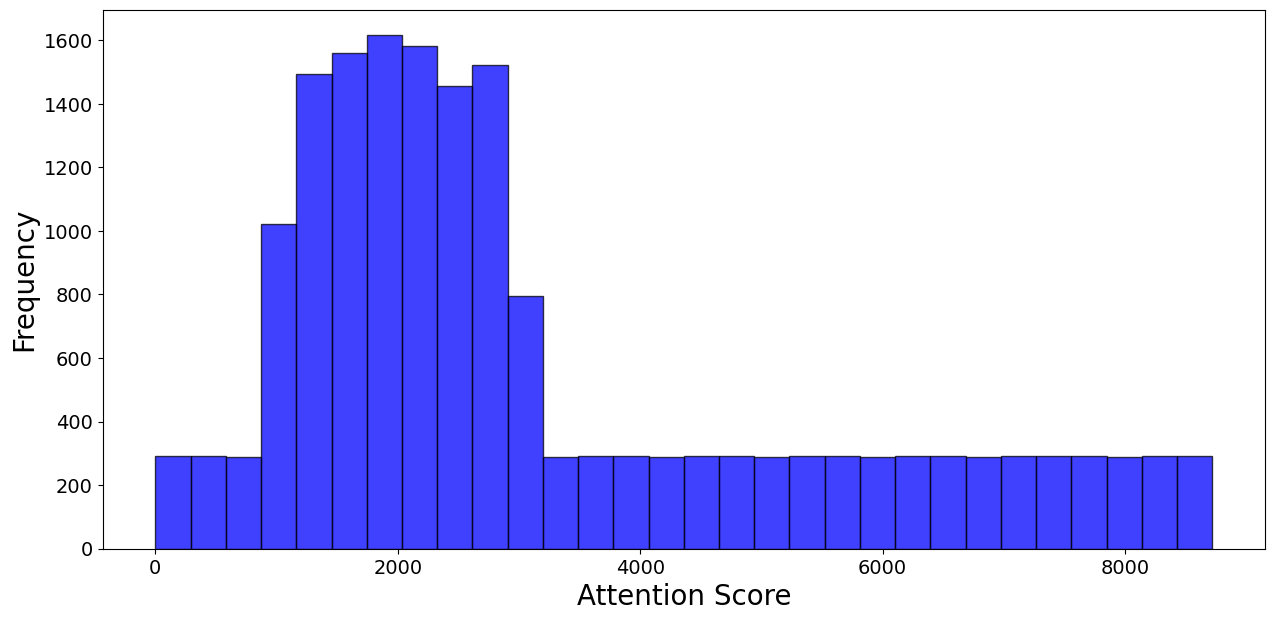}
    \caption{Attention weights to the climate interactions}
\end{figure}
The x-axis in the histogram represents the attention scores, and the y-axis represents the frequency of nodes receiving those scores. The peak and distribution highlight our model's focus areas, with most nodes receiving moderate attention and a select few receiving very high attention. The distribution shows a long tail extending towards higher attention scores. 

Next,  we conduct a systemtic feature ablation study to further interpret the model. Specifically, we remove one feature at a time (e.g., sentiment scores, textual embeddings, etc.) and observe how the model's performance changes, which provides an idea of which features are most important for the model.
\begin{table*}[!htb]
    \centering
    \caption{Feature Ablation Study Results. Notes: (1) CSGAT: ClimateSent-GAT with all features. (2) No Par Emb: Model without parent embeddings. (3) No Ch Emb: Model without child embeddings. (4) No Par Sent: Model without parent entity-based sentiment scores. (5) No Ch Sent: Model without child entity-based sentiment scores.}
    \label{tab:ablation_results}
    \begin{tabular}{lcccccc}
        \hline
        Metric / Ablated Feature & CSGAT & No Par Emb & No Ch Emb & No Par Sent & No Ch Sent \\
        \hline
        \hline
        Accuracy & \textbf{0.79} & 0.61 & 0.56 & 0.69 & \underline{0.70} \\
        Macro Avg F1 & \textbf{0.78} & 0.59 & 0.54 & 0.67 & \underline{0.69} \\
        Disagree F1 & \textbf{0.82} & 0.68 & 0.61 & 0.75 & \underline{0.74} \\
        Neutral F1 & \textbf{0.72} & 0.47 & 0.48 & 0.60 & \underline{0.63} \\
        Agree F1 & \textbf{0.79} & 0.62 & 0.54 & 0.67 & \underline{0.68} \\
        Disagree Precision & \textbf{0.87} & 0.74 & 0.68 & 0.78 & \underline{0.78} \\
        Neutral Precision & \textbf{0.65} & 0.40 & 0.43 & 0.55 & \underline{0.58} \\
        Agree Precision & \textbf{0.78} & 0.61 & 0.52 & 0.66 & \underline{0.68} \\
        Disagree Recall & \textbf{0.78} & 0.62 & 0.56 & 0.72 & \underline{0.71} \\
        Neutral Recall & \textbf{0.81} & 0.57 & 0.54 & 0.66 & \underline{0.71} \\
        Agree Recall & \textbf{0.80} & 0.63 & 0.56 & 0.67 & \underline{0.68} \\
        \hline
    \end{tabular}
\end{table*}

The ablation study shows that all of our features contribute information, so that omitting any one of them impairs performance. Replicating what \citet{PougueBiyong2023} report for BERT-base classification across all five DEBAGREEMENT subreddits,  the "Without Child Embeddings" condition yields the worst performance.  The child comments are reactions to parent comments, and appear to provide more specific, task-relevant, information about whether the interaction is an agreement or a disagreement. 

Omitting either parent entity-based sentiment or child-entity-based sentiment impairs performance, but surprisingly, the ablated model that omits child entity-based sentiment performs slightly better.  This is counterintuitive and might warrant further investigation. Possibly, the child's sentiment is somewhat redundant with other features, especially if the textual embeddings of child comments are already rich in sentiment information.

\begin{figure}[h]
    \centering
    \includegraphics[width=0.9\columnwidth]{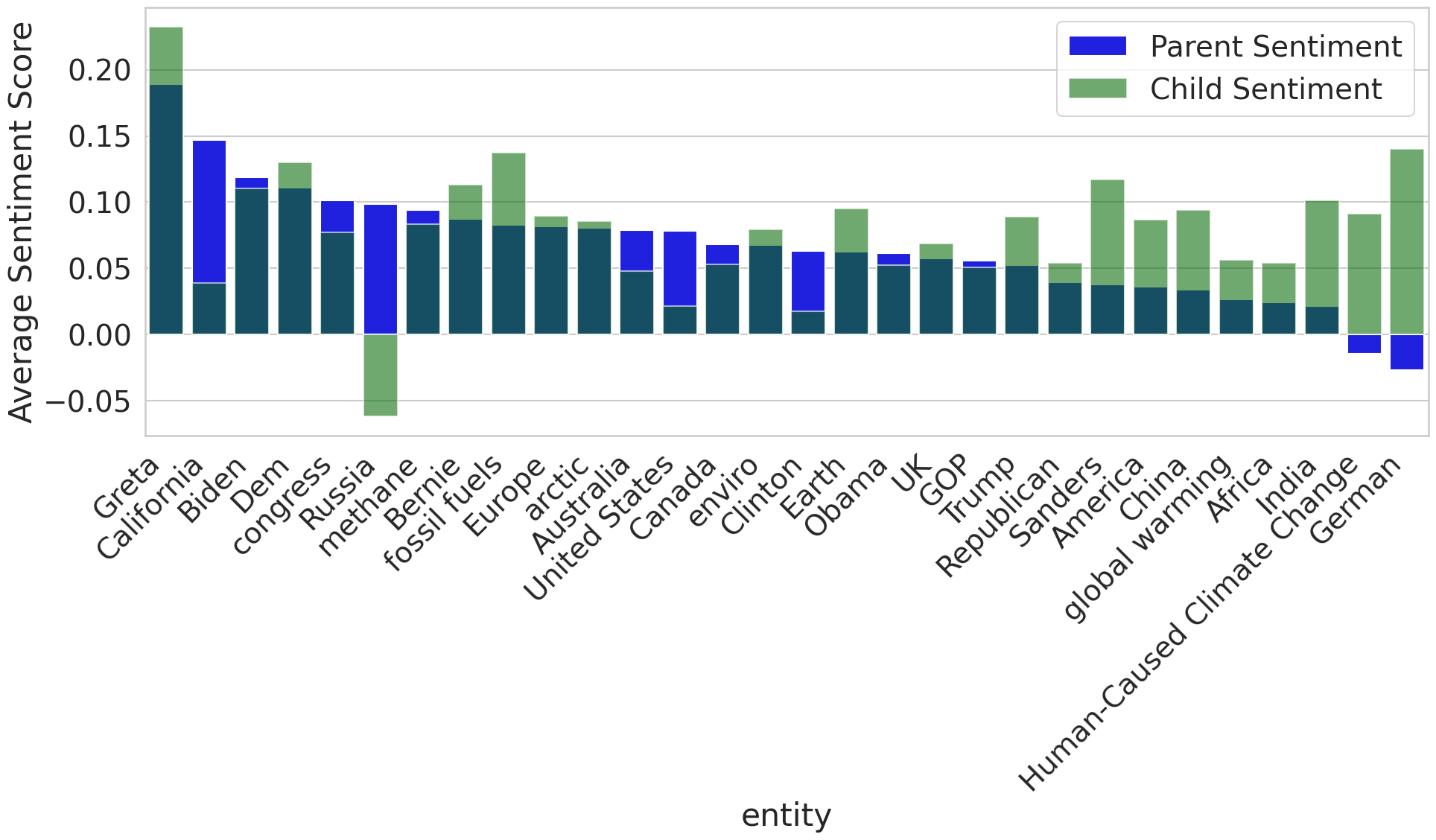}
    \caption{Parent and Child Sentiment by Climate-Related Entities}
\end{figure}

\begin{figure*}[h]
    \centering
    \includegraphics[width=0.6\textwidth]{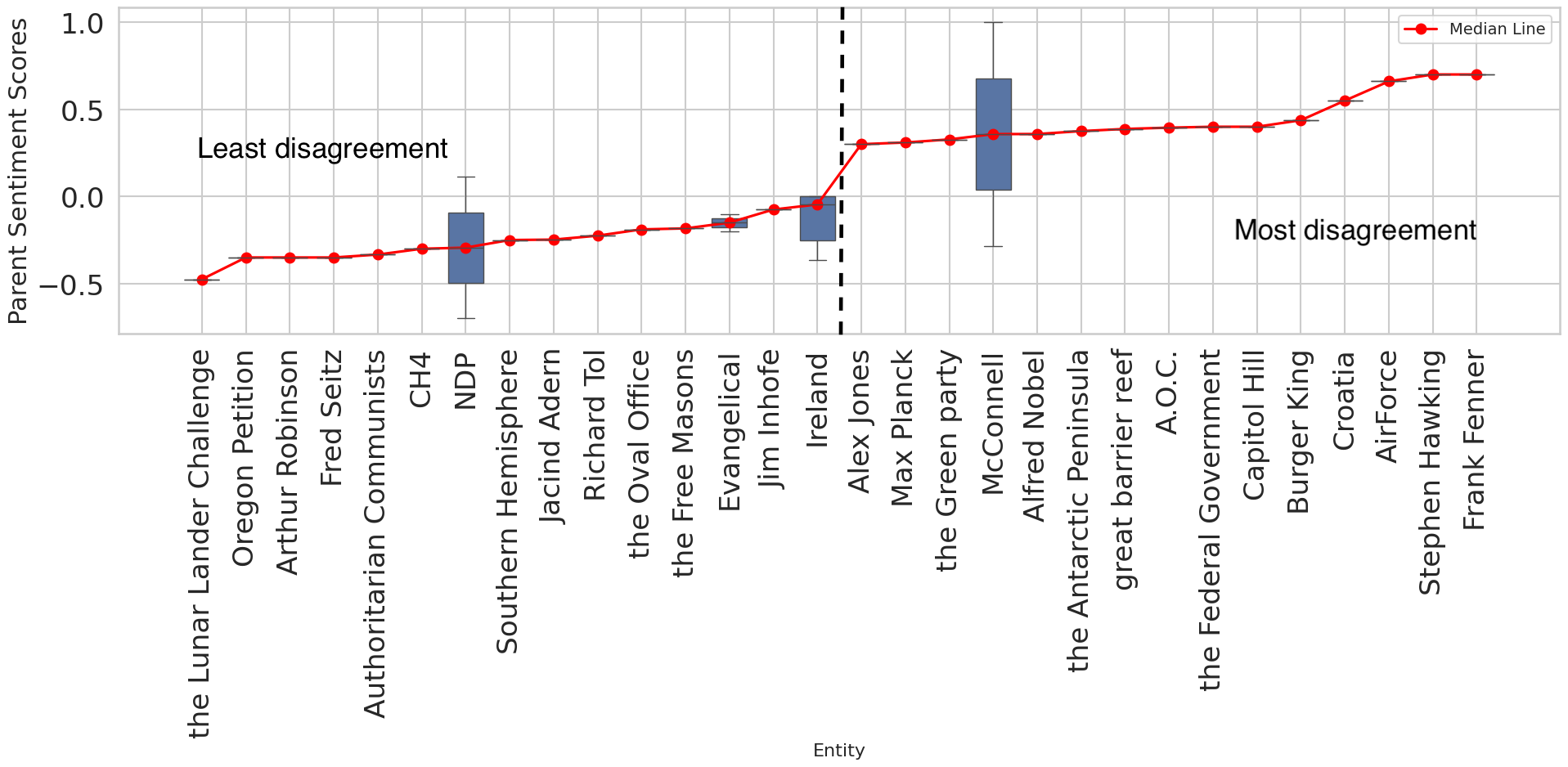}
    \caption{Parent sentiment for entities with least vs most disagreement }
\end{figure*}


In conclusion, the feature ablation studies help in understanding the importance of different feature sets in the model. These can be interpreted as an indication of how conversational context or sentiment may affect the model's ability to classify social media (dis)agreements. 

The performance of ClimateSent-GAT underscores the value of incorporating both graph attention mechanisms and robust pre-trained language models in understanding complex social interactions on social media platforms in the climate change discourse. It holds promise for real-world applications for disagreement detection as well.

\subsubsection{Climate Entities-Based Analysis}
Our methodology makes it possible to identify specific factors and issues relating to (dis)agreements in on-line discourse about climate. 

To investigate how (dis)agreements are shaped around climate-related entities, we selected the 30 most frequently occurring entities to visualize their average entity-based sentiment scores and label distributions. Figure 3 illustrates the varying degrees of sentiment between parent and child comments across most frequent-discussed entities, such as "Greta," "California," and "Trump." Generally, child comments exhibit less negative sentiments compared to their parent counterparts. This trend may suggest that child comments often serve to counteract the tone set by parent comments. 

Figure 4 compares the parent sentiment for the entities mentioned in the least versus the most disagreements. Entities on the x-axis are sorted by disagreement percentages in ascending order. Notably, entities involved in more disagreements tend to have higher median parent sentiment scores. This pattern may indicate that when a parent user refers to a climate entity in a more positive manner, the child user often presents a contrasting opinion. We also observed that there is no apparent correlation between sentiment differences and levels of disagreement. 

These observations are indicative of the complex interplay between the sentiment expressed and the class of (dis)agreements in the comments. Such dynamics are crucial for understanding how public opinions on climate issues are shaped and propagated through social media platforms. Please see the Appendix for a complete form of (dis)agreements distributions and entity-based sentiment scores.

\begin{figure*}[h]
    \centering
    \includegraphics[width=0.55\textwidth]{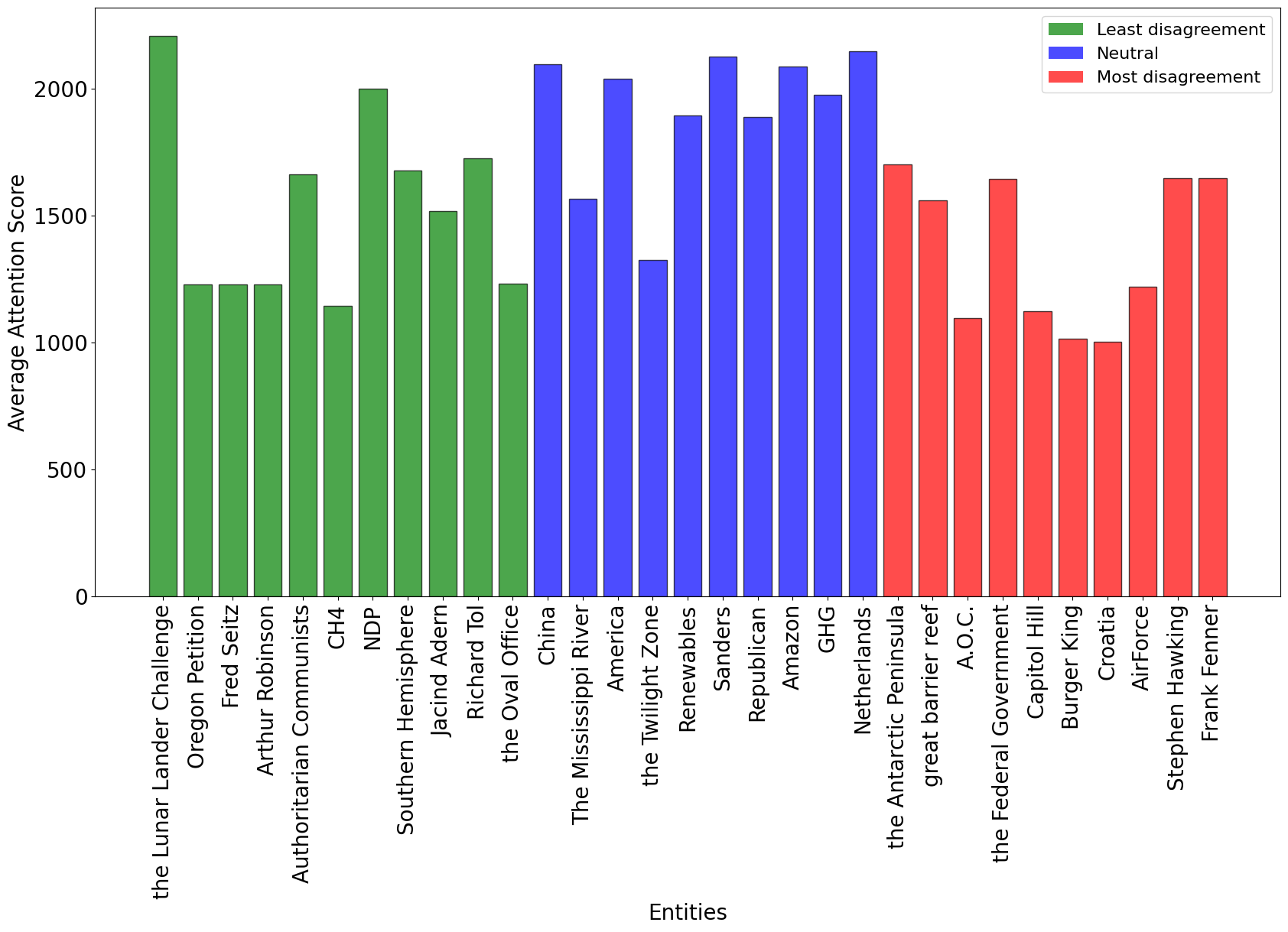}
    \caption{Average Sentiment Attention Scores for Different Entity Categories }
\end{figure*}

To analyze the features or labels of nodes receiving the highest attention scores, we selected the entities associated with the most disagreeing/agreeing/neutral interactions, and extracted the relevant attention weights. The most entities most associated with disagreement include 'Frank Fenner', 'Stephen Hawking', 'AirForce', 'Croatia', 'Burger King', 'Capitol Hill', 'the Federal Government', 'A.O.C.', 'great barrier reef', and 'the Antarctic Peninsula'; the entities associated with most agreement include 'Richard Tol', 'Jacind Adern', 'Southern Hemisphere', 'NDP', 'CH4', 'Authoritarian Communists', 'Arthur Robinson', 'Fred Seitz', 'Oregon Petition', and 'the Lunar Lander Challenge'; the most neutral ones include 'Netherlands', 'GHG', 'Amazon', 'Republican', 'Sanders', 'Renewables', 'the Twilight Zone', 'America', 'The Mississippi River', and 'China'.


The differences in attention displayed in Figure 5 support our suggestion that the ability to learn entity-specific attention weights is a factor in the success of our model. Overall, entities associated with neutral discussions receive the highest attention, followed by entities associated with agreement, and finally entities associated with disagreement. These results mirror the fact that the neutral category is the most difficult one to classify, followed by the agreement category, followed by the disagreement category. We note that the neutral category combines several different sorts of discourse; some neutral posts are so-classified because they do not have any language that expresses strong attitudes, while others have been classified as neutral because they agree in some respects while disagreement in others. It appears that paying attention to the specific entities under discussion helps the model to navigate the nuances of communication in these cases.

\section{Related Work}

Several notable works precede ours in using graph-based approaches on the DEBAGREEMENT dataset.  \citet{PougueBiyong2023} use a community-detection algorithm on social networks defined by the interactions, in order to compare the dynamics of polarization in different subreddit communities. 
\citet{Lorge2024} successfully predicted disagreements in a slice of the DEBAGREEMENT dataset, using a Signed Graph Convolutional Network (SGCN) applied to a a bipartite graph organized around the stance of users towards selected named entities. This study showcases the ability of GNNs to discern subtleties in discourse relations that traditional models often miss. Our ClimateSent-GAT achieves better generality by leveraging not only the structural data but also more extensive textual information.  We employ Graph Attention Networks (GATs) instead of other graph neural networks, because of their dynamic edge weighting and fine-grained attention capabilities. These features enable the model to adaptively focus on the most relevant interactions within social media discussions, which is crucial for accurately detecting disagreements and integrating diverse data types such as textual embeddings and sentiment scores.

\section{Limitations}

We acknowledge several limitations that future research could address. Firstly, the reliance on data from a single social media platform (Reddit) might limit the generalizability of the model. Social media platforms vary significantly in user demographics and interaction styles, which can influence discourse dynamics and the manifestation of disagreement.  Secondly, the inherent complexity of the Graph Attention Network (GAT) architecture used in our model could pose challenges in terms of interpretability and computational demands, which may limit deployment when scaling up the model. Lastly, while we have taken significant steps to address ethical considerations, particularly concerning data privacy and the potential for misuse of disagreement detection technologies, these remain critical ongoing concerns. Future iterations of this research should consider multi-platform studies, enhanced methods for handling linguistic nuances, and further exploration of ethical implications in the deployment of NLP technologies in climate discourse.

\section*{Acknowledgements}
RS is deeply grateful for the support provided by the Jardine Scholarship, which has made this research possible. JBP was supported by the Engineering and Physical Sciences Research Council (EP/T023333/1).

\section*{Appendix A: Label Distribution by Climate-Related Entities}
Below is the table showing the percentage distribution of labels (Agree, Disagree, Neutral) for various climate-related entities, we manually sampled 176 entities which contain figures, geographic locations, institutions, climate topics and agreements as they are of most importance when studying the (dis)agreements of climate discourse on online platforms.

\onecolumn
\appendix
\section*{Appendix A: Label Distribution by Climate-Related Entities}

\begin{longtable}{@{}p{0.4\textwidth}rrr@{}}
\caption{Percentage distribution of agreement labels across different climate-related entities.} \label{tab:long} \\
\toprule
\textbf{Entity} & \textbf{Agree (\%)} & \textbf{Disagree (\%)} & \textbf{Neutral (\%)} \\
\midrule
\endfirsthead
\multicolumn{4}{c}%
{{\bfseries Table \thetable\ continued from previous page}} \\
\toprule
\textbf{Entity} & \textbf{Agree (\%)} & \textbf{Disagree (\%)} & \textbf{Neutral (\%)} \\
\midrule
\endhead
\bottomrule
\endfoot

A.O.C. & 100 & 0 & 0 \\
ACB & 100 & 0 & 0 \\
Africa & 30 & 55 & 15 \\
AirForce & 0 & 100 & 0 \\
Al gore & 0 & 100 & 0 \\
Alex Jones & 0 & 100 & 0 \\
Alfred Nobel & 100 & 0 & 0 \\
Amazon & 38.46 & 46.15 & 15.38 \\
America & 34.02 & 51.55 & 14.43 \\
Antarctic & 23.08 & 69.23 & 7.69 \\
Arthur Robinson & 100 & 0 & 0 \\
Asia & 33.33 & 41.67 & 25 \\
Augsburg University & 100 & 0 & 0 \\
Australia & 52.78 & 30.56 & 16.67 \\
Authoritarian Communists & 100 & 0 & 0 \\
Baltimore & 100 & 0 & 0 \\
Bernie & 27.78 & 55.56 & 16.67 \\
Biden & 27.5 & 52.5 & 20 \\
Bill Nye & 0 & 100 & 0 \\
Bitcoin & 14.29 & 85.71 & 0 \\
Brexit & 100 & 0 & 0 \\
Bruce Willis & 0 & 100 & 0 \\
Bundesverband WindEnergie & 100 & 0 & 0 \\
Burger King & 100 & 0 & 0 \\
CERN & 0 & 100 & 0 \\
CH4 & 0 & 100 & 0 \\
California & 20 & 65 & 15 \\
Canada & 27.27 & 45.45 & 27.27 \\
Capitalism & 7.69 & 92.31 & 0 \\
Capitol Hill & 0 & 100 & 0 \\
Chevron & 0 & 0 & 100 \\
China & 27.45 & 54.9 & 17.65 \\
Clinton & 44.44 & 50 & 5.56 \\
Conservative & 33.33 & 50 & 16.67 \\
Coronavirus & 50 & 0 & 50 \\
Croatia & 0 & 0 & 100 \\
Cube Satellites & 0 & 100 & 0 \\
Dem & 38.89 & 44.44 & 16.67 \\
ESA & 0 & 100 & 0 \\
EU & 38.46 & 38.46 & 23.08 \\
Earth & 27.78 & 58.33 & 13.89 \\
El-Nino & 25 & 75 & 0 \\
Elon & 40 & 60 & 0 \\
Environmental Genocide & 0 & 100 & 0 \\
Europe & 31.25 & 43.75 & 25 \\
Evangelical & 0 & 100 & 0 \\
Exxon & 30.77 & 53.85 & 15.38 \\
Faux News & 100 & 0 & 0 \\
Finland & 100 & 0 & 0 \\
Florida & 58.33 & 16.67 & 25 \\
Frank Fenner & 0 & 100 & 0 \\
Fred Seitz & 100 & 0 & 0 \\
GBR & 0 & 100 & 0 \\
GE & 33.33 & 66.67 & 0 \\
GHG & 20 & 70 & 10 \\
GOP & 33.33 & 45.83 & 20.83 \\
Georgetown & 0 & 100 & 0 \\
German & 35.71 & 50 & 14.29 \\
Gibson & 0 & 50 & 50 \\
Great Lakes & 50 & 0 & 50 \\
Green New Deal & 14.29 & 57.14 & 28.57 \\
Greenpeace & 66.67 & 33.33 & 0 \\
Greg James & 100 & 0 & 0 \\
Greta & 55.56 & 33.33 & 11.11 \\
Gwynne Dyer & 0 & 0 & 100 \\
Halifax & 100 & 0 & 0 \\
Harvey & 25 & 75 & 0 \\
Heartland Institute & 50 & 0 & 50 \\
Hillary & 53.85 & 38.46 & 7.69 \\
Holly Gillibrand & 0 & 100 & 0 \\
Hollywood & 100 & 0 & 0 \\
Holochain & 0 & 100 & 0 \\
Human-Caused Climate Change & 21.43 & 71.43 & 7.14 \\
IEA & 100 & 0 & 0 \\
IPCC & 30.77 & 46.15 & 23.08 \\
India & 25 & 46.88 & 28.12 \\
Inslee & 36.36 & 45.45 & 18.18 \\
Ireland & 40 & 20 & 40 \\
Israel & 0 & 0 & 100 \\
Italy & 50 & 50 & 0 \\
Jacind Adern & 100 & 0 & 0 \\
Japan & 55.56 & 22.22 & 22.22 \\
Jim Inhofe & 0 & 100 & 0 \\
Kardashev & 0 & 100 & 0 \\
Kevin Anderson & 0 & 100 & 0 \\
KristophMcKane & 0 & 100 & 0 \\
LNG & 100 & 0 & 0 \\
La Nina & 100 & 0 & 0 \\
Leonardo DiCaprio & 100 & 0 & 0 \\
Liberal & 0 & 66.67 & 33.33 \\
Lithium & 0 & 50 & 50 \\
Mark Zuckerberg & 50 & 50 & 0 \\
Mars & 60 & 40 & 0 \\
Max Planck & 0 & 0 & 100 \\
McConnell & 50 & 50 & 0 \\
McPherson & 0 & 80 & 20 \\
Miami Beach & 100 & 0 & 0 \\
Michael McCabe & 0 & 100 & 0 \\
Michale Bays Armageddon & 0 & 100 & 0 \\
Myron Ebell & 100 & 0 & 0 \\
NAFTA & 100 & 0 & 0 \\
NASA & 16.67 & 83.33 & 0 \\
NATO & 100 & 0 & 0 \\
NDP & 0 & 100 & 0 \\
NOAA & 20 & 50 & 30 \\
Naomi Klein & 33.33 & 33.33 & 33.33 \\
Nature Communications & 100 & 0 & 0 \\
Netherlands & 33.33 & 33.33 & 33.33 \\
New Zealand & 50 & 50 & 0 \\
North Hemisphere & 0 & 0 & 100 \\
Norway & 60 & 40 & 0 \\
Obama & 17.24 & 62.07 & 20.69 \\
Ohio & 33.33 & 33.33 & 33.33 \\
Oregon Petition & 100 & 0 & 0 \\
PBS & 100 & 0 & 0 \\
PURE CO2 & 0 & 0 & 100 \\
Phoenicians & 100 & 0 & 0 \\
Pocahontas & 0 & 100 & 0 \\
PricewaterhouseCoopers & 0 & 0 & 100 \\
Propaganda & 0 & 50 & 50 \\
Renewables & 0 & 100 & 0 \\
Republican & 33.96 & 50.94 & 15.09 \\
Richard Tol & 0 & 100 & 0 \\
Royal Dutch Shell & 0 & 0 & 100 \\
Russia & 35 & 45 & 20 \\
Sanders & 29.41 & 52.94 & 17.65 \\
Saudi Arabia & 0 & 100 & 0 \\
Scotland & 0 & 50 & 50 \\
Silicon & 0 & 100 & 0 \\
Socialism & 0 & 100 & 0 \\
Solar & 16.67 & 66.67 & 16.67 \\
South Korea & 100 & 0 & 0 \\
Southern Hemisphere & 0 & 100 & 0 \\
Stephen Hawking & 0 & 100 & 0 \\
Switzerland & 50 & 0 & 50 \\
The BC Liberals & 0 & 100 & 0 \\
The Mississippi River & 100 & 0 & 0 \\
The Paris Agreement & 0 & 0 & 100 \\
The Relative Sea Level of the Sargasso Sea & 0 & 0 & 100 \\
Thunberg & 66.67 & 16.67 & 16.67 \\
Trudeau & 50 & 25 & 25 \\
Trump & 39.64 & 43.24 & 17.12 \\
Tucson & 50 & 0 & 50 \\
UK & 46.67 & 40 & 13.33 \\
United States & 56.25 & 25 & 18.75 \\
VP Gore & 100 & 0 & 0 \\
Warren & 25 & 66.67 & 8.33 \\
Western Europe India & 0 & 0 & 100 \\
White House & 25 & 50 & 25 \\
YouTube & 57.14 & 28.57 & 14.29 \\
arctic & 22.58 & 61.29 & 16.13 \\
christian & 33.33 & 66.67 & 0 \\
citizens climate lobby & 100 & 0 & 0 \\
congress & 29.41 & 64.71 & 5.88 \\
ecosia & 0 & 66.67 & 33.33 \\
enviro & 35.14 & 47.97 & 16.89 \\
fossil fuels & 23.29 & 67.12 & 9.59 \\
global warming & 25.81 & 56.45 & 17.74 \\
great barrier reef & 100 & 0 & 0 \\
green new deal & 33.33 & 33.33 & 33.33 \\
healthcare & 50 & 50 & 0 \\
methane & 33.33 & 39.39 & 27.27 \\
the Antarctic Peninsula & 0 & 100 & 0 \\
the Federal Government & 0 & 100 & 0 \\
the Free Masons & 0 & 0 & 100 \\
the Green party & 100 & 0 & 0 \\
the Holocene Extinction & 0 & 0 & 100 \\
the Koch Brothers & 100 & 0 & 0 \\
the Lunar Lander Challenge & 100 & 0 & 0 \\
the New York Times & 100 & 0 & 0 \\
the Oval Office & 0 & 100 & 0 \\
the Planetary Society & 0 & 100 & 0 \\
the Supreme Court & 100 & 0 & 0 \\
the Twilight Zone & 100 & 0 & 0 \\
the Washington Post & 100 & 0 & 0 \\
zero hours & 0 & 100 & 0 \\

\end{longtable}
\twocolumn

\onecolumn
\section*{Appendix B: Sentiment Analysis of Climate-Related Entities}

Below is the table presents the average parent and child sentiment scores for various climate-related entities identified in social media discussions. The entities are sorted by percentage of 'Disagree' in descending order.

\begin{longtable}{@{}p{0.4\textwidth}rrr@{}}
\caption{Parent and child sentiment scores for climate-related entities.} \label{tab:long} \\
\toprule
\hline
\textbf{Entity} & \textbf{Parent Sentiment} & \textbf{Child Sentiment} \\
\hline
\midrule
\endfirsthead
\multicolumn{4}{c}%
{{\bfseries Table \thetable\ continued from previous page}} \\
\toprule
\hline
\textbf{Entity} & \textbf{Parent Sentiment} & \textbf{Child Sentiment} \\
\hline
\midrule
\endhead
\bottomrule
\endfoot
Frank Fenner & 0.7 & 0.1375 \\
Stephen Hawking & 0.7 & 0.1375 \\
AirForce & 0.6604166667 & 0.19 \\
Croatia & 0.55 & 0.1637662338 \\
Burger King & 0.5375 & 0 \\
Capitol Hill & 0.5 & 0.2393939394 \\
the Federal Government & 0.4 & 0.1583333333 \\
A.O.C. & 0.3947916667 & 0.25 \\
great barrier reef & 0.3875 & 0.4333333333 \\
the Antarctic Peninsula & 0.375 & 0.1 \\
Alfred Nobel & 0.3583333333 & 0 \\
McConnell & 0.3578125 & 0.1145833334 \\
the Green party & 0.3272727273 & -0.225 \\
Max Planck & 0.3095454545 & -0.2291666667 \\
Alex Jones & 0.3 & 0 \\
Pocahontas & 0.2888888889 & -0.002083333335 \\
Thunberg & 0.2867063492 & 0.3514814815 \\
Brexit & 0.2857142857 & 0.08333333333 \\
the Planetary Society & 0.2843537415 & 0.1201388889 \\
Cube Satellites & 0.28 & -0.08333333333 \\
Georgetown & 0.2579166667 & 0.15 \\
Chevron & 0.25 & 0.15 \\
Kardashev & 0.25 & 0 \\
Israel & 0.2380952381 & 0.25 \\
Western Europe India & 0.2380952381 & 0.25 \\
Gwynne Dyer & 0.2277777778 & 0.01285714286 \\
LNG & 0.225 & -0.228125 \\
CERN & 0.2166666667 & -0.3125 \\
ESA & 0.2166666667 & -0.3125 \\
NAFTA & 0.2144444444 & 0.039375 \\
Environmental Genocide & 0.2 & 0 \\
The Paris Agreement & 0.2 & -0.08273809524 \\
healthcare & 0.1971064815 & 0.1478174603 \\
Great Lakes & 0.1907061688 & 0.2111111111 \\
ACB & 0.19 & 0.1333333333 \\
the Supreme Court & 0.19 & 0.1333333333 \\
Greta & 0.1892405203 & 0.232546162 \\
PBS & 0.1891836735 & 0.5590909091 \\
KristophMcKane & 0.1875 & 0.2333333333 \\
Holly Gillibrand & 0.1833333333 & 0.06277056277 \\
Mars & 0.1795833333 & 0.3049206349 \\
Halifax & 0.1666666667 & -0.08928571429 \\
Finland & 0.1636363637 & -0.06916666667 \\
Warren & 0.1528736772 & 0.05767609127 \\
Ohio & 0.1518253968 & -0.01598639456 \\
Antarctic & 0.1477039627 & 0.07841783217 \\
California & 0.1470361652 & 0.03901541081 \\
Augsburg University & 0.1465909091 & 0.475 \\
Nature Communications & 0.1465909091 & 0.475 \\
Faux News & 0.14 & -0.1666666667 \\
Florida & 0.1391583243 & 0.06483503596 \\
ecosia & 0.1361342593 & -0.06572420635 \\
IEA & 0.1315277778 & 0.1779761905 \\
the New York Times & 0.1308001894 & -0.1958333334 \\
Gibson & 0.1291666667 & 0.2158333333 \\
El-Nino & 0.1276271645 & 0.04766253093 \\
Socialism & 0.1267361111 & 0.233030303 \\
Saudi Arabia & 0.1266067266 & 0.2 \\
Royal Dutch Shell & 0.125 & 0.1810606061 \\
Myron Ebell & 0.1242897727 & -0.55 \\
Inslee & 0.1224621212 & 0.03557900433 \\
Mark Zuckerberg & 0.1210961657 & 0.09044642859 \\
Biden & 0.1190405318 & 0.1103702946 \\
Switzerland & 0.1170833334 & 0.0773809524 \\
The BC Liberals & 0.1125 & 0 \\
Dem & 0.1110492462 & 0.1301153817 \\
Japan & 0.1094157848 & 0.07711940837 \\
Liberal & 0.1091550926 & -0.02612433862 \\
Kevin Anderson & 0.1071712018 & 0.02777777778 \\
Conservative & 0.1066633598 & 0.05432249078 \\
Solar & 0.1064361472 & 0.05602141955 \\
congress & 0.101569448 & 0.07718646549 \\
Russia & 0.09845155424 & -0.06192766955 \\
YouTube & 0.09619897959 & 0.1575633031 \\
La Nina & 0.09444444444 & 0.475 \\
Hillary & 0.09399343711 & 0.03209917859 \\
methane & 0.09377795815 & 0.08348263934 \\
Norway & 0.09333333334 & -0.036 \\
christian & 0.09 & -0.07354497356 \\
Baltimore & 0.08944444445 & 0.2061011904 \\
Bernie & 0.08729056437 & 0.1133162645 \\
citizens climate lobby & 0.08518518519 & 0.65 \\
fossil fuels & 0.08275026586 & 0.1372720437 \\
Europe & 0.08151242927 & 0.08970922253 \\
arctic & 0.08057866685 & 0.08545000699 \\
Exxon & 0.07927655678 & 0.1033248696 \\
Australia & 0.07911194883 & 0.04762864258 \\
GE & 0.07882689744 & 0.253131905 \\
The Relative Sea Level of the Sargasso Sea & 0.07857142857 & 0.09761904762 \\
United States & 0.07839781746 & 0.02161907017 \\
EU & 0.07350434822 & 0.02837598115 \\
Bundesverband WindEnergie & 0.07012987013 & 0.2583333333 \\
Asia & 0.06996527777 & 0.1062872024 \\
Canada & 0.06830349399 & 0.05283802309 \\
enviro & 0.06743364128 & 0.07928235043 \\
Scotland & 0.0662037037 & 0.0427412518 \\
Elon & 0.06619444444 & 0.08861111111 \\
Clinton & 0.06288359788 & 0.01759749779 \\
Naomi Klein & 0.06284722223 & 0.06041666668 \\
Earth & 0.06251891916 & 0.09536423694 \\
Obama & 0.06153171182 & 0.05222946593 \\
UK & 0.05744136375 & 0.0688579771 \\
Miami Beach & 0.05648148148 & -0.3 \\
VP Gore & 0.05648148148 & -0.3 \\
GOP & 0.05583540014 & 0.0507129162 \\
Trump & 0.05260003078 & 0.0890837389 \\
Netherlands & 0.04428571429 & -0.02833333333 \\
GHG & 0.04258547008 & 0.07759920635 \\
Amazon & 0.04178747179 & 0.04293402112 \\
Republican & 0.03949567035 & 0.05396672248 \\
Sanders & 0.03760270775 & 0.1168629785 \\
Renewables & 0.03757936509 & 0.0442770713 \\
the Twilight Zone & 0.03666666667 & -0.0625 \\
America & 0.03620889243 & 0.08652358673 \\
The Mississippi River & 0.03617424242 & 0.2888888889 \\
China & 0.03377156548 & 0.09396245761 \\
the Washington Post & 0.03166666667 & -0.1702020202 \\
White House & 0.03088624339 & 0.07337729978 \\
NASA & 0.02848260096 & 0.1450578704 \\
Coronavirus & 0.02732954546 & 0.08479166667 \\
New Zealand & 0.02711715366 & 0.2658820347 \\
global warming & 0.02663607786 & 0.05606813624 \\
IPCC & 0.02529853479 & -0.01513680763 \\
Africa & 0.02424829001 & 0.05433479368 \\
India & 0.0215511114 & 0.1012734551 \\
Michale Bays Armageddon & 0.02142857143 & 0.08474358974 \\
Green New Deal & 0.0145302614 & 0.1220716089 \\
the Koch Brothers & 0.0047222222 & 0.08125 \\
Propaganda & 0.003571428572 & -0.02395833334 \\
Bill Nye & 0.002324263033 & 0.01226851853 \\
the Holocene Extinction & 0 & 0.5 \\
zero hours & 0 & 0.3333333333 \\
Hollywood & 0 & 0.2787878788 \\
Phoenicians & 0 & 0.2787878788 \\
Bruce Willis & 0 & 0.2266666667 \\
Michael McCabe & 0 & 0.2240909091 \\
PricewaterhouseCoopers & 0 & 0.2121212121 \\
North Hemisphere & 0 & -0.0625 \\
Al gore & 0 & -0.1333333334 \\
NATO & 0 & -0.2 \\
Holochain & 0 & -0.28125 \\
GBR & 0 & -0.3777777778 \\
South Korea & -4.63E-18 & 0.274702381 \\
Lithium & -0.001666666665 & 0.1489795918 \\
Greg James & -0.002857142857 & -0.25 \\
Leonardo DiCaprio & -0.01111111111 & -0.1166666667 \\
Tucson & -0.01166666667 & -0.2106060606 \\
Human-Caused Climate Change & -0.01432539683 & 0.09102419406 \\
Trudeau & -0.01696180555 & 0.00535714286 \\
Italy & -0.01944444445 & -0.01 \\
Capitalism & -0.02398203647 & 0.07151251526 \\
Bitcoin & -0.02544075965 & 0.1215455576 \\
Greenpeace & -0.02575757575 & 0.1690972222 \\
German & -0.02703836342 & 0.1403067666 \\
Heartland Institute & -0.0277777778 & 0.15 \\
PURE CO2 & -0.0287202381 & -0.1875 \\
green new deal & -0.03453102453 & 0.06912878788 \\
NOAA & -0.04918741733 & 0.03659722222 \\
Silicon & -0.05333333333 & -0.2020408163 \\
Harvey & -0.05892857143 & 0.1897321429 \\
McPherson & -0.06264646463 & 0.03785714286 \\
Jim Inhofe & -0.075 & -0.15 \\
Ireland & -0.1325 & 0.1344642857 \\
Evangelical & -0.15 & -0.193030303 \\
the Free Masons & -0.1833333333 & -0.125 \\
the Oval Office & -0.1888888889 & 0.2107142857 \\
Richard Tol & -0.225 & 0.1510416667 \\
Jacind Adern & -0.2475 & -0.05555555556 \\
Southern Hemisphere & -0.25 & -0.175 \\
NDP & -0.29375 & 0.0681818182 \\
CH4 & -0.3 & 0.5 \\
Authoritarian Communists & -0.3333333333 & -0.1333333333 \\
Arthur Robinson & -0.35 & 0.5 \\
Fred Seitz & -0.35 & 0.5 \\
Oregon Petition & -0.35 & 0.5 \\
the Lunar Lander Challenge & -0.475 & 0.2875 \\
\end{longtable}
\twocolumn

\nocite{*}

\bibliography{acl2023}
\bibliographystyle{acl_natbib}

\end{document}